\documentclass{article}
\usepackage{arxiv} %
\usepackage{graphicx} %
\usepackage{amsmath}
\usepackage{amsthm}
\usepackage{amssymb}
\usepackage{xcolor}
\usepackage{xspace}
\usepackage{booktabs}
\usepackage{longtable}
\usepackage{makecell}
\usepackage[hidelinks]{hyperref}
\usepackage{xurl} %
\usepackage{subcaption}
\usepackage{amsmath, amssymb}
\usepackage{float}
\usepackage{placeins} %
\usepackage{mathtools}
\usepackage{thmtools} %

\makeatletter
\DeclareRobustCommand\onedot{\futurelet\@let@token\@onedot}
\def\@onedot{\ifx\@let@token.\else.\null\fi\xspace}

\def\eg{e.g\onedot}
\def\ie{i.e\onedot}

\makeatother

\newcommand{\vu}{{\boldsymbol{u}}}
\newcommand{\vv}{{\boldsymbol{v}}}
\newcommand{\vw}{{\boldsymbol{w}}}
\newcommand{\vpsi}{{\boldsymbol{\psi}}}

\newcommand{\vrho}{{\boldsymbol{\rho}}}

\newcommand{\vxi}{{\boldsymbol{\xi}}}
\newcommand{\vsigma}{{\boldsymbol{\sigma}}}
\newcommand{\mA}{{\mathcal{A}}}
\newcommand{\mB}{{\mathcal{B}}}
\newcommand{\mU}{{\mathcal{U}}}
\newcommand{\mV}{{\mathcal{V}}}
\newcommand{\mW}{{\mathcal{W}}}

\newcommand{\mQ}{{\mathcal{Q}}}

\renewcommand{\S}{{\mathbb{S}}}
\newcommand{\R}{{\mathbb{R}}}

\newcommand{\N}{{\mathbb{N}}}

\newcommand{\lambdainv}{{\lambda^{-1}}}
\newcommand{\range}[1]{[#1]}

\newcommand{\set}[1]{\left\{ #1 \right\}}

\newcommand{\id}{I}
\newcommand{\vid}{\boldsymbol{\id}}

\newcommand{\mset}[1]{\{\!\!\{ #1 \}\!\!\}}
\newcommand{\multisets}[1]{\mathcal{M}(#1)}

\newcommand{\stackedtensor}[1]{\left[ \begin{matrix}#1\end{matrix} \right]}
\newcommand{\tupledtensor}[1]{\left\langle \begin{matrix}#1\end{matrix} \right\rangle}
\newcommand{\openinterval}[1]{{]}#1{[}}

\makeatletter
\newcommand{\lift}[1]{\lift@check#1\relax\@nil{#1}}
\def\lift@check#1#2\@nil#3{%
  \ifx\relax#2\bar{#3}\else\overline{#3}\fi}
\makeatother
\newcommand{\optensor}[3]{#3 \circ \lift{#2} \circ #1}
\newcommand{\omegatensor}[2]{\optensor{#1}{#2}{\Omega}}
\newcommand{\sumtensor}[2]{\optensor{#1}{#2}{\Sigma}}
\newcommand{\prodtensor}[2]{\optensor{#1}{#2}{\Pi}}
\newcommand{\maxtensor}[2]{\optensor{#1}{#2}{\vee}}

\newcommand{\einsumtensor}[2]{\sumtensor{#1^{-1}}{\prodtensor{\vid}{\stackedtensor{#2}}}}

\newcommand{\fun}[3]{{#1}:{#2}\to{#3}}
\DeclareMathOperator{\supp}{supp}

\newcommand{\thinrule}{\specialrule{0.005cm}{0.1cm}{0.1cm}}
\newcommand{\smallbullet}{\hspace{0.65cm}}

\theoremstyle{definition}
\newtheorem{definition}{Definition}

\newtheorem{lemma}{Lemma}

\title{Representing Tensors as Compositions for Efficient Computation on Sparse and Structured Data}
\renewcommand{\shorttitle}{Representing Tensors as Compositions}
\author{Val\'erian Rey \\ SimplexLab \\ \texttt{valerian.rey@gmail.com} \And Praneth Paruchuri \\ SimplexLab \\ \texttt{pranethparuchuri@gmail.com}}
\date{}

\begin{document}

\maketitle

\begin{abstract}
    We propose a new mathematical formalism in which all kinds of tensors and of operations on those tensors are represented as simple function compositions. Tensors are functions from indices to values. An element-wise operation (e.g. an activation function) is a composition with a function applied to values; re-indexing (e.g. reshape, transposition) is a composition with a function applied to indices; and a reduction is a composition with both an index function, mapping each index to several indices, and a reduction operation (e.g. product, sum, max). Combined with tensor stacking, these compositions can represent most interesting kinds of tensors (e.g. dense, jagged, all kinds of sparsity patterns, matrix factorizations), as well as lazy versions of almost every operation available in modern tensor computation frameworks, including einsums between tensors of any kind.
    Identities between compositions can then be used to rewrite such expressions into equivalent ones, without evaluating them. For instance, the product of two sparse matrices can be rewritten so that only products between non-zero values are computed, turning a sparse problem into a dense one.
    Using this algebra as the intermediate representation of a tensor computation engine would shift the focus from defining operations on tensors (e.g. einsum between sparse tensors) to defining an algebra on index functions. This should reduce the codebase complexity from $\mathcal{O}(mn^2)$, with $m$ the number of binary operations between tensors and $n$ the number of tensor types, to $\mathcal{O}(m + k^2)$, with $k$ the number of types of index functions. A draft implementation of a tensor computation framework based on this algebra is publicly available.\footnote{\raggedright \mbox{\url{https://github.com/SimplexLab/cten}}}
\end{abstract}

\section{Introduction}

In many tensor computation frameworks, such as \texttt{numpy} and \texttt{PyTorch}, tensors are represented as dense arrays, with some limited support for alternative tensor types, such as the sparse format \texttt{sparse\_coo} or jagged (nested) tensors. One of the reasons for such limited support is that each unary operation has to be defined for all tensor types, and each binary operation has to be defined for all possible pairs of tensor types (e.g. if you want to make a matrix product between a matrix in sparse COO format, and another one in sparse CSR format, you would need an implementation specifically for that). This makes the complexity of the codebase grow quadratically with the number of tensor types.

This combinatorial problem is well known in sparse compilation, which dates back to the work of Bik~\cite{bik1996thesis}. TACO~\cite{kjolstad2017taco} describes the format of each operand as a composition of per-dimension level formats sharing a common interface~\cite{chou2018format}. The sparse tensor dialect of MLIR~\cite{bik2022mlir} adopts such per-dimension level types to annotate tensor types, and compiles operations written without any knowledge of sparsity into sparse code.
The Universal Sparse Tensor~\cite{bik2026ust} goes further by describing formats as mappings from tensor dimensions to storage levels.
It then either dispatches the operations to optimized libraries when the format is usual, or generates code otherwise.
Perhaps the closest work to ours is StructTensor~\cite{ghorbani2023structtensor}, which infers the sparsity and redundancy of tensors symbolically at compile time, and its follow-up DASTAC~\cite{ghorbani2025dastac}, which also packs structured tensors into dense buffers.
These systems address the problem with compiler and dispatch machinery, whereas we address it algebraically; their code generation techniques are
complementary to our work.

In this paper, we propose an alternative approach that unifies how tensors and operations on tensors are represented.
It relies on representing tensors as mappings from indices to values. Operations can then be represented using compositions. This can represent a very large number of operations, including element-wise operations (e.g. activation functions), re-indexings (e.g. reshapes, transpositions), reductions (e.g. summing or multiplying over a dimension) as well as composite operations like einsums (e.g. matrix multiplications, tensor products, convolutions, attention). Surprisingly, this can also directly be used to represent any pattern of sparsity (e.g. coordinate (COO) format, block diagonal) as well as jagged tensors, and, by composition, any of the operations mentioned above between any kind of sparse or jagged tensor. Since each function can freely be composed with any other functions as long as the domains match, this mathematical framework is extremely modular. Its implementation, despite being non-trivial, should scale very well to a large number of operations and sparsity formats.

\section{Theory}
\label{sec:theory}

Throughout this paper, we use the notation $\range{k}$ for $\set{1, \dots, k}$. Elements of a Cartesian product are written as tuples, \eg $(m, n) \in \range{M} \times \range{N}$, and a $1$-tuple $(i)$ is identified with $i$.
We denote by $\circ$ the standard function composition: for $f: \mV \to \mU$ and $g: \mU \to \mW$, $g \circ f: \mV \to \mW$ is defined as $(g \circ f)(\vv) = g(f(\vv))$.

\subsection{Tensors}

\begin{definition}[Tensor]
    Let $\mU$ be a finite set, called an \emph{index set}, and let $\S$ be a set. The function $U: \mU \to \S$ is a \emph{tensor} on $\mU$.
\end{definition}

A vector in $\R^n$ can be represented by a tensor $\fun{V}{\range{n}}\R$. Similarly, a matrix in $\R^{m\times n}$ can be represented by a tensor $\fun{M}{\range{m}\times\range{n}}\R$.
Since tensors are functions, they can be composed with other functions. Composing a tensor $U: \mU \to \S$ with a function on values $f: \S \to \S'$ gives the tensor $f \circ U: \mU \to \S'$, which applies $f$ element-wise. Composing it with a function on indices $\rho: \mV \to \mU$ gives the tensor $U \circ \rho: \mV \to \S$, which re-indexes $U$: each index of the result is mapped to the index of $U$ at which its value is read.

\subsection{Multisets and index functions}

A \emph{multiset} is an unordered collection in which elements may appear more than
once. Formally, a multiset of size $n$ over a set $\S$ is an equivalence
class of $n$-tuples in $\S^n$ under permutation of coordinates; we write
$\mset{x_1,\dots,x_n}$ for the class of $(x_1,\dots,x_n)$. Thus
$\mset{a,a,b} = \mset{a,b,a} \neq \mset{a,b}$. Further, we denote by $\multisets{\S}$ the set of multisets of any size over $\S$, including the empty multiset.
The \emph{sum} of two multisets is the multiset containing the elements of both, with multiplicities added:
\[
  \mset{x_1, \dots, x_n} \uplus \mset{y_1, \dots, y_m} = \mset{x_1, \dots, x_n, y_1, \dots, y_m}.
\]

Let $f : \mV \to \mU$ be a function. We define its extension to multisets
$\bar{f} : \multisets{\mV} \to \multisets{\mU}$ as
\[
  \bar{f}\left(\mset{v_1, \dots, v_n}\right) = \mset{f(v_1), \dots, f(v_n)}.
\]

Intuitively, $\bar{f}$ is the function that applies $f$ to each element of its input multisets.
We call a function $\psi : \mV \to \multisets{\mU}$ an \emph{index function}
from $\mV$ to $\mU$: it maps each index in $\mV$ to any number of indices in $\mU$. We identify a set with the multiset in which each of
its elements appears exactly once. For any set $\mathcal{X}$, we denote by
$\eta : \mathcal{X} \to \multisets{\mathcal{X}}$ the \emph{singleton} map,
defined as $\eta(x) = \mset{x}$.
We call a function $\rho : \mV \to \mU$ an \emph{index map} from $\mV$ to $\mU$: it maps each index in $\mV$ to exactly one index in $\mU$. An index map $\rho$ thus gives the index function $\eta \circ \rho$, which maps each $\vv$ to $\mset{\rho(\vv)}$. The \emph{inverse} of an index map
$\rho : \mU \to \mV$ is the index function $\rho^{-1} : \mV \to \multisets{\mU}$
mapping each $\vv \in \mV$ to its preimage $\set{\vu \in \mU ~\middle|~ \rho(\vu) = \vv}$.

\begin{definition}[Support and intersection of index functions]
    Let $\psi : \mV \to \multisets{\mU}$ be an index function. The \emph{support} of $\psi$ is the set of indices that $\psi$ maps to a non-empty multiset:
    \begin{equation}
        \supp \psi = \set{\vv \in \mV ~\middle|~ \psi(\vv) \neq \mset{}}.
    \end{equation}
    Let $k \in \N$, and for each $i \in \range{k}$, let $\psi_i : \mV \to \multisets{\mU_i}$ be an index function. By a slight abuse of notation, we write:
    \begin{equation}
        \psi_1 \cap \dots \cap \psi_k = \supp \psi_1 \cap \dots \cap \supp \psi_k.
    \end{equation}
\end{definition}

\subsection{Reductions}

We define a \emph{reduction} on a set $\S$ as a function $\Omega : \multisets{\S} \to \S$ for which there exists $e \in \S$, called its \emph{neutral element}, such that, for all $x \in \S$ and all $X, Y \in \multisets{\S}$:
\begin{align*}
    \Omega(\mset{}) &= e, \\
    \Omega(\mset{x}) &= x, \\
    \Omega(X \uplus Y) &= \Omega\left(\mset{\Omega(X), \Omega(Y)}\right).
\end{align*}
The first two properties say that reducing the empty multiset gives the neutral element, and that reducing a single element gives this element. The third one is associativity: a multiset can be reduced by splitting it into two parts, reducing each part independently, and then reducing the two results. Commutativity is built into the definition: since multisets are unordered, $\Omega$ cannot depend on the order of the elements it reduces.
Equivalently, $\Omega$ combines the elements of a multiset, in any order, with the associative and commutative binary operation $(x, y) \mapsto \Omega(\mset{x, y})$, which has $e$ as neutral element.
For instance, the sum $\Sigma$ and the product $\Pi$ are reductions on $\R$, with neutral elements $0$ and $1$, and the maximum $\vee$ and the minimum $\wedge$ are reductions on $\R \cup \set{-\infty, \infty}$, with neutral elements $-\infty$ and $\infty$.

\begin{definition}[Reduction tensor]
    Let $U: \mU \to \S$ be a tensor, let $\psi: \mV \to \multisets{\mU}$ be an index function, and let $\Omega$ be a reduction on $\S$. The \emph{reduction tensor} with \emph{underlying tensor} $U$ and \emph{index function} $\psi$ is the composition
    \begin{equation}
        \omegatensor{\psi}{U}: \mV \to \S,
    \end{equation}
    which maps each $\vv \in \mV$ to the reduction of the multiset $\lift{U}(\psi(\vv))$ of the values of $U$ at the indices of $\psi(\vv)$.
\end{definition}

When $\Omega$ is the sum $\Sigma$, we call it a \emph{sum tensor}, and similarly for other reductions. A reduction tensor is equal to the neutral element $e$ of $\Omega$ outside of $\supp \psi$.

\begin{definition}[Sparse tensor]
\label{def:sparse}
    Let $\psi : \mV \to \multisets{\mU}$ be an index function mapping every index to a multiset containing at most one element, and let $U : \mU \to \R$ be a tensor. The \emph{sparse tensor} with \emph{underlying tensor} $U$ and \emph{index function} $\psi$ is the sum tensor
    \begin{equation}
        \sumtensor{\psi}{U} : \mV \to \R.
    \end{equation}
\end{definition}

A sparse tensor is thus zero outside of $\supp \psi$: only its values on $\supp \psi$ are stored, in $U$, and $\psi$ tells where they are. Most sparse formats, such as COO without duplicates, satisfy the condition that each index is mapped to at most one stored value.

\subsection{Stacks}

\begin{definition}[Stack]
    Let $k \in \N$, and for each $i \in \range{k}$, let $U_i : \mU_i \to \S$ be a tensor. Finally, let $\mU=\bigcup_{i \in \range{k}} \{i\} \times \mU_i$.
    The \emph{stack}
    $\stackedtensor{U_1 & \cdots & U_k}: \mU \to \S$
    is defined as:
    \begin{equation}
        \stackedtensor{U_1 & \cdots & U_k}\quad:\quad (i, \vu)\in \mU \quad\mapsto\quad U_i(\vu).
    \end{equation}
\end{definition}

Note that the index sets $\mU_i$ need not be equal: we can stack tensors of different shapes. For instance, stacking vectors $V_i: \range{n_i} \to \R$ of different lengths $n_i$ yields a jagged tensor $\stackedtensor{V_1 & \cdots & V_k}: \bigcup_{i \in \range{k}} \set{i} \times \range{n_i} \to \R$.

When dealing with stacks, it is very useful to be able to combine several index functions (one per tensor in the stack) into an index function that can be used with the entire stack. This is defined as follows.

\begin{definition}[Stacked index function]
    Let $k \in \N$, let $\mV$ be an index set, and for each $i \in \range{k}$, let
    $\psi_i : \mV \to \multisets{\mU_i}$ be an index function.
    The \emph{stacked index function}
    $\vpsi : \mV \to \multisets{\bigcup_{i \in \range{k}} \{i\} \times \mU_i}$
    is defined as:
    \begin{equation}
        \vpsi(\vv) = \mset{(1, \vu_{1,1}), \dots, (1, \vu_{1,n_1})} \uplus \dots \uplus \mset{(k, \vu_{k,1}), \dots, (k, \vu_{k,n_k})}
    \end{equation}
    where $\psi_i(\vv) = \mset{\vu_{i,1}, \dots, \vu_{i,n_i}}$ for each $i \in \range{k}$.
    We extend this to index maps: for each $i \in \range{k}$, let $\rho_i : \mV \to \mU_i$ be an index map. Their \emph{stacked index function} $\vrho : \mV \to \multisets{\bigcup_{i \in \range{k}} \{i\} \times \mU_i}$ is defined as:
    \begin{equation}
        \vrho(\vv) = \mset{(1, \rho_1(\vv)), \dots, (k, \rho_k(\vv))}.
    \end{equation}
    In particular, when the $\rho_i$ are all the identity function on $\mV$, we call $\vrho$ the \emph{stacked identity} and denote it by $\vid$:
    \begin{equation}
        \vid(\vv) = \mset{(1, \vv), \dots, (k, \vv)}.
    \end{equation}
\end{definition}

\section{Examples}
\label{sec:examples}

We now illustrate how common tensors and operations are represented in our algebra, from the simplest to the most composite. The translation of many common PyTorch operations is given in Appendix~\ref{app:translations}. More complex operations, such as the layers of transformers, convolutions and vectorization (\texttt{vmap}), can also be expressed in our algebra. Code snippets use PyTorch, where \texttt{F} denotes \texttt{torch.nn.functional}.

\begin{figure}[tbp]
  \centering
  \begin{subfigure}{\linewidth}
    \centering
    \includegraphics[width=0.64\linewidth]{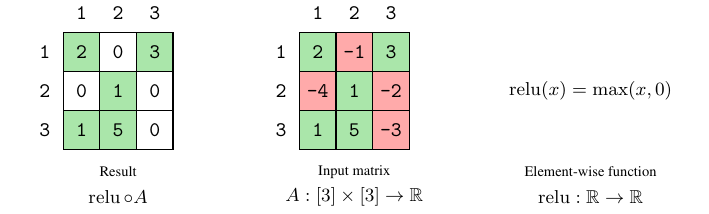}
    \caption{Element-wise ReLU (\texttt{torch.relu(A)}). Negative entries (red) are mapped to $0$.}
    \label{fig:relu}
  \end{subfigure}

  \vspace{0.25em}

  \begin{subfigure}{\linewidth}
    \centering
    \includegraphics[width=0.64\linewidth]{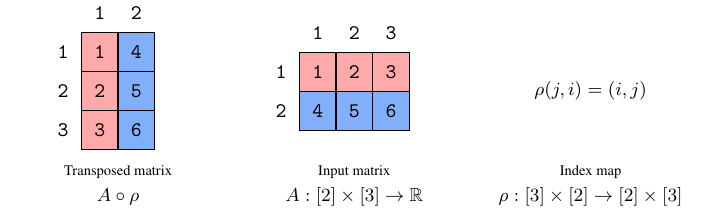}
    \caption{Transposition of a matrix (\texttt{A.T}). The index map swaps the two indices.}
    \label{fig:transpose}
  \end{subfigure}

  \vspace{0.25em}

  \begin{subfigure}{\linewidth}
    \centering
    \includegraphics[width=0.64\linewidth]{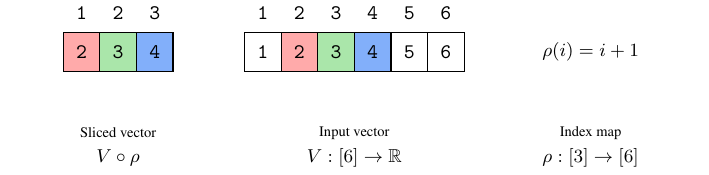}
    \caption{Slicing of a vector (\texttt{V[1:4]}). The index map shifts the indices.}
    \label{fig:slice}
  \end{subfigure}

  \vspace{0.25em}

  \begin{subfigure}{\linewidth}
    \centering
    \includegraphics[width=0.64\linewidth]{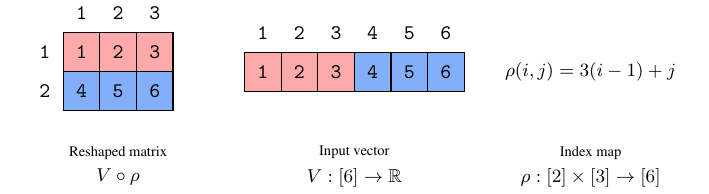}
    \caption{Reshape of a vector into a matrix (\texttt{V.reshape(2, 3)}).}
    \label{fig:reshape}
  \end{subfigure}

  \vspace{0.25em}

  \begin{subfigure}{\linewidth}
    \centering
    \includegraphics[width=0.64\linewidth]{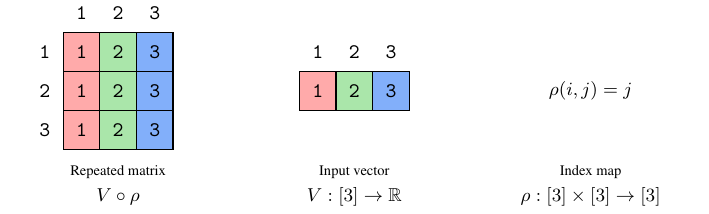}
    \caption{Repetition of a vector (\texttt{V.repeat(3, 1)}). The index map ignores the row index.}
    \label{fig:repeat}
  \end{subfigure}

  \caption{Compositions with a function: on values for element-wise operations (a), and on indices for re-indexings (b--e).}
  \label{fig:compositions}
\end{figure}

\paragraph{Compositions with a function.}
The simplest operations compose a tensor with a single function (Figure~\ref{fig:compositions}): a function on values for element-wise operations, as in $\mathrm{relu} \circ A$, or an index map for re-indexings, as in $A \circ \rho$. The index map need not be injective: repeating a vector maps several indices of the result to the same index of the input.

\begin{figure}[tbp]
  \centering
  \begin{subfigure}{\linewidth}
    \centering
    \includegraphics[width=0.74\linewidth]{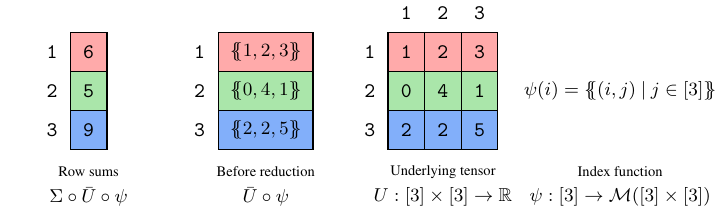}
    \caption{Sum of a matrix over its second axis (\texttt{torch.sum(U, dim=1)}). The index function maps each row index to the indices of this row.}
    \label{fig:sum-axis}
  \end{subfigure}

  \vspace{0.25em}

  \begin{subfigure}{\linewidth}
    \centering
    \includegraphics[width=0.74\linewidth]{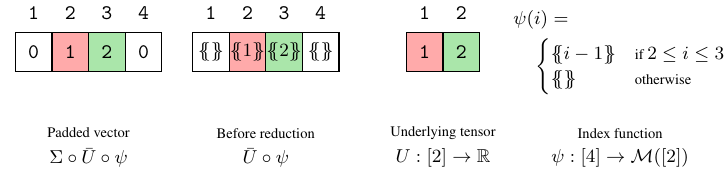}
    \caption{Zero-padding of a vector (\texttt{F.pad(U, (1, 1))}). Padded entries are mapped to the empty multiset, whose sum is $0$, so these zeros do not use any memory in this representation.}
    \label{fig:pad}
  \end{subfigure}

  \vspace{0.25em}

  \begin{subfigure}{\linewidth}
    \centering
    \includegraphics[width=0.74\linewidth]{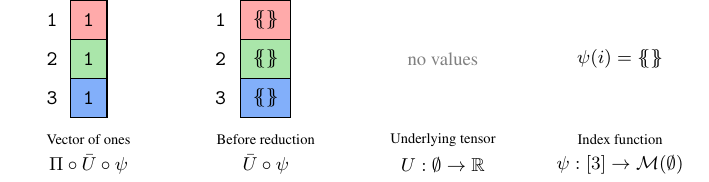}
    \caption{Vector of ones (\texttt{torch.ones(3)}). No data has to be stored at all.}
    \label{fig:ones}
  \end{subfigure}

  \caption{Reduction tensors. The index function maps each index to a multiset of indices of the underlying tensor, whose values, gathered by $\bar{U} \circ \psi$, are then reduced.}
  \label{fig:reductions}
\end{figure}

\paragraph{Reductions.}
Figure~\ref{fig:reductions} shows reduction tensors $\Omega \circ \bar{U} \circ \psi$. Summing a matrix over an axis maps each row index to all the indices of this row. Since a reduction tensor is equal to the neutral element outside of the support of its index function (Section~\ref{sec:theory}), mapping indices to the empty multiset gives zero-padding with a sum, and a vector of ones with a product, without storing any value.

\begin{figure}[tbp]
  \centering
  \begin{subfigure}{\linewidth}
    \centering
    \includegraphics[width=0.74\linewidth]{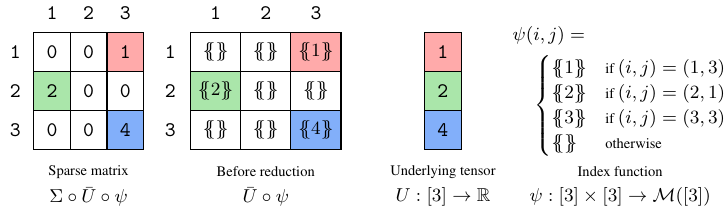}
    \caption{COO matrix. The non-zero values and their coordinates (in the representation of $\psi$) are stored.}
    \label{fig:coo}
  \end{subfigure}

  \vspace{0.25em}

  \begin{subfigure}{\linewidth}
    \centering
    \includegraphics[width=0.74\linewidth]{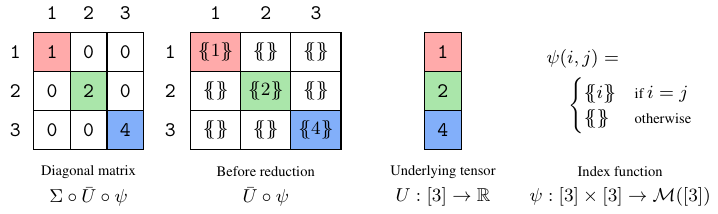}
    \caption{Diagonal matrix. Only the diagonal values have to be stored. The representation of $\psi$ does not hold any data.}
    \label{fig:diag}
  \end{subfigure}

  \caption{Sparse matrices, represented as sum tensors.}
  \label{fig:sparse}
\end{figure}

\paragraph{Sparse tensors.}
A sparse tensor $\Sigma \circ \bar{U} \circ \psi$ (Definition~\ref{def:sparse}) only stores its non-zero values, in the underlying tensor $U$, and the index function $\psi$ tells where they are (Figure~\ref{fig:sparse}). In the COO format, $\psi$ is the inverse of the function mapping the index in $U$ of each stored value to its coordinates, which are stored explicitly. For structured sparsity patterns, such as diagonal matrices, $\psi$ is given by a formula, so the coordinates need not be stored at all. Diagonal matrices can of course be combined with the vector of ones of Figure~\ref{fig:ones}: taking it as the underlying tensor of the diagonal matrix of Figure~\ref{fig:diag} gives the identity matrix, without needing any memory.

\begin{figure}[tbp]
  \centering
  \begin{subfigure}{\linewidth}
    \centering
    \includegraphics[width=0.74\linewidth]{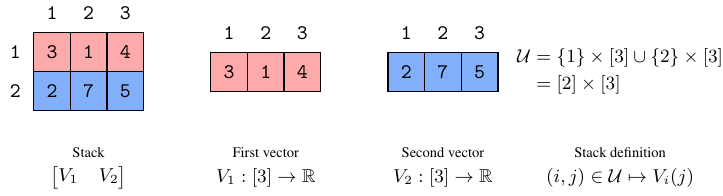}
    \caption{Stack of two vectors (\texttt{torch.stack([V1, V2])}).}
    \label{fig:stack}
  \end{subfigure}

  \vspace{0.25em}

  \begin{subfigure}{\linewidth}
    \centering
    \includegraphics[width=0.74\linewidth]{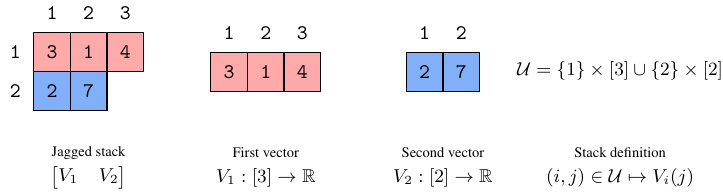}
    \caption{Stack of two vectors of different lengths, giving a jagged tensor. Unlike in (a), the stacked tensors have different index sets.}
    \label{fig:jagged}
  \end{subfigure}

  \vspace{0.25em}

  \begin{subfigure}{\linewidth}
    \centering
    \includegraphics[width=0.74\linewidth]{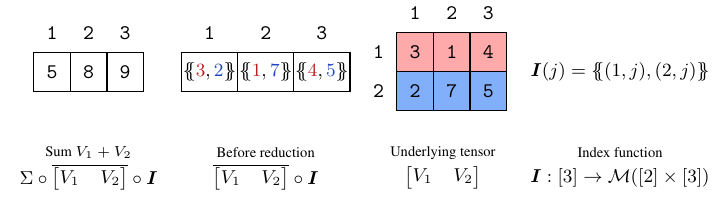}
    \caption{Element-wise addition of the two vectors of (a) (\texttt{V1 + V2}). The stacked identity $\vid$ gathers the values of $V_1$ and $V_2$ at each index, which are then summed.}
    \label{fig:add}
  \end{subfigure}

  \caption{Stacking, and element-wise operations between several tensors, represented as reductions over a stack.}
  \label{fig:stacking}
\end{figure}

\paragraph{Operations on several tensors.}
Several tensors can be combined by stacking them into a single tensor (Figure~\ref{fig:stacking}). An element-wise operation between them, such as an addition, is then a reduction over the stack, whose index function gathers the values of all operands at the same index: this is the stacked identity $\vid$.

\begin{figure}[tbp]
  \centering
  \begin{subfigure}{\linewidth}
    \centering
    \includegraphics[width=0.956\linewidth]{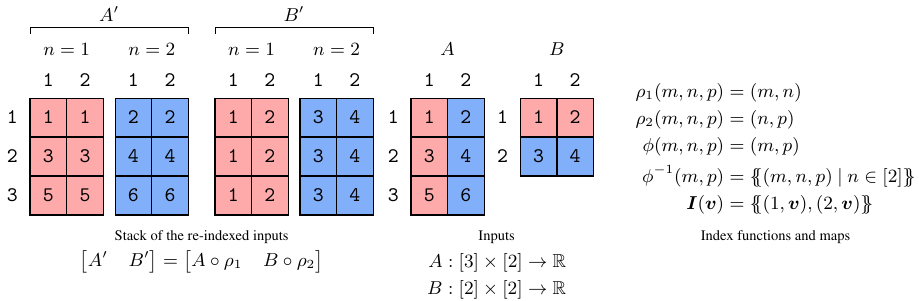}
    \caption{Re-indexing: the index maps $\rho_1$ and $\rho_2$ bring $A$ and $B$ to the same latent index set $\range{3} \times \range{2} \times \range{2}$, giving $A' = A \circ \rho_1$ and $B' = B \circ \rho_2$, which can be multiplied element-wise.}
    \label{fig:matmul-stack}
  \end{subfigure}

  \vspace{0.25em}

  \begin{subfigure}{\linewidth}
    \centering
    \includegraphics[width=0.956\linewidth]{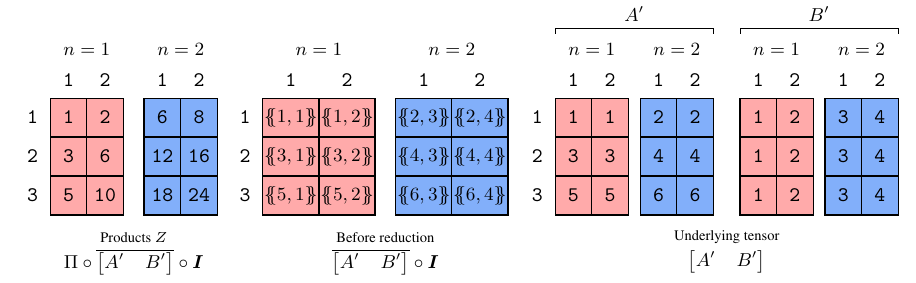}
    \caption{Product: for each latent index $(m, n, p)$, the stacked identity $\vid$ gathers the multiset $\mset{A(m, n), B(n, p)}$, whose product is $Z(m, n, p)$.}
    \label{fig:matmul-prod}
  \end{subfigure}

  \vspace{0.25em}

  \begin{subfigure}{\linewidth}
    \centering
    \includegraphics[width=0.956\linewidth]{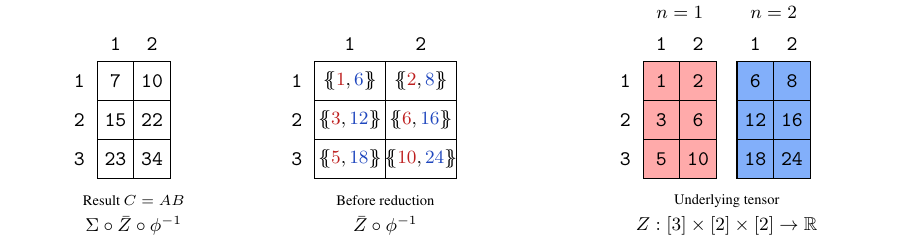}
    \caption{Sum: for each output index $(m, p)$, $\phi^{-1}$ selects the products $Z(m, n, p)$ of all latent indices $(m, n, p)$, which are summed.}
    \label{fig:matmul-sum}
  \end{subfigure}

  \caption{Dense matrix multiplication $C = AB$, represented as an einsum. In each subfigure, the result is the underlying tensor of the next one. This illustrates the definition, not an actual implementation.}
  \label{fig:matmul}
\end{figure}

\paragraph{Einsums.}
Combining all of the above, an einsum between several tensors first re-indexes each operand to a common latent index set, then reduces their stack with a product, and finally sums the products over the latent indices that are not in the output. For instance, the matrix multiplication $C = AB$ is represented as
\begin{equation*}
    C = \sumtensor{\phi^{-1}}{\prodtensor{\vid}{\stackedtensor{A \circ \rho_1 & B \circ \rho_2}}},
\end{equation*}
with the index maps and index functions given in Figure~\ref{fig:matmul-stack}. This is illustrated in Figure~\ref{fig:matmul}.

\section{Algebraic rewriting}
\label{sec:rewriting}

Since tensors and operations on tensors are compositions of functions, identities between compositions can be used to rewrite an expression into an equivalent one, before evaluating it. This is particularly useful for sparse tensors. Evaluated directly from its definition, an operation on sparse tensors goes through every index of its domain, including the many indices at which its values are zero. In this section, we show that common operations on sparse tensors can be rewritten, using only identities between compositions, into sparse tensors whose underlying tensors are computed from the stored values of the operands only:
\begin{itemize}
    \item a zero-preserving element-wise function, such as $\mathrm{relu}$ or $\tanh$, only has to be applied to the stored values (Theorem~\ref{thm:sparse-ew});
    \item the element-wise product of sparse tensors only requires the products at the indices of the intersection of their supports (Theorem~\ref{thm:sparse-prod});
    \item more generally, an einsum between sparse tensors only requires the products of stored values at the latent indices at which all operands have a stored value, which are then summed for each output index (Theorem~\ref{thm:sparse-einsum}).
\end{itemize}
In all of these rewritings, the only non-trivial computations are on index functions: the support of the result, given by the intersection of the supports of the operands, an enumeration of this support, and compositions of index functions. Values are only accessed through the underlying tensors. The rewritings are thus independent of the sparse format: the same identities hold whatever the index functions of the operands, and an implementation only has to compute intersections and compositions of index functions efficiently, based on their structure, instead of defining each operation for each combination of formats.
The proofs, as well as the general identities between compositions on which they rely, are given in Appendix~\ref{app:proofs}.

\begin{restatable}[Zero-preserving element-wise function]{theorem}{thmsparseew}
\label{thm:sparse-ew}
    Let $A = \sumtensor{\psi}{U}: \mV \to \R$ be a sparse tensor, and let $f: \R \to \R$ be a zero-preserving function, \ie $f(0) = 0$. Then:
    \begin{equation}
        f \circ \sumtensor{\psi}{U} = \sumtensor{\psi}{f \circ U}.
        \label{eq:sparse-ew}
    \end{equation}
\end{restatable}

This means that a zero-preserving element-wise function can be applied to only the stored values of the sparse tensor.

\begin{restatable}[Element-wise product of sparse tensors]{theorem}{thmsparseprod}
\label{thm:sparse-prod}
    For each $i \in \range{k}$, let $A_i = \sumtensor{\psi_i}{U_i}: \mV \to \R$ be a sparse tensor, with $\psi_i: \mV \to \multisets{\mU_i}$ and $U_i: \mU_i \to \R$. Let $\mQ = \psi_1 \cap \dots \cap \psi_k$, of size $L$, and let $\lambda: \range{L} \to \mV$ be an injective function with image $\mQ$ (\eg enumerating $\mQ$ in lexicographic order). Then:
    \begin{equation}
        \prodtensor{\vid}{\stackedtensor{A_1 & \cdots & A_k}} = \sumtensor{\lambdainv}{\prodtensor{\vsigma}{\stackedtensor{U_1 & \cdots & U_k}}},
        \label{eq:sparse-prod}
    \end{equation}
    where $\vsigma$ is the stacked index function of $\sigma_i = \psi_i \circ \lambda: \range{L} \to \multisets{\mU_i}$, for $i \in \range{k}$.
\end{restatable}

\begin{figure}[tbp]
  \centering
  \begin{subfigure}{\linewidth}
    \centering
    \includegraphics[width=0.826\linewidth]{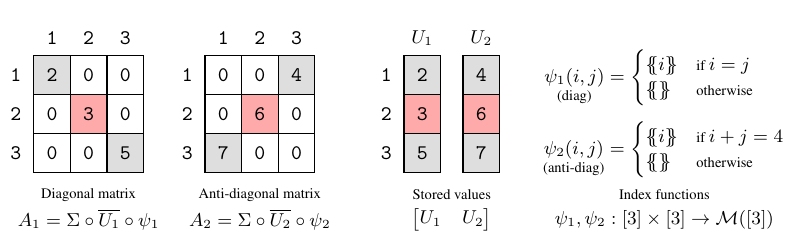}
    \caption{The sparse inputs: a diagonal matrix $A_1$ and an anti-diagonal matrix $A_2$, with their stored values $U_1$ and $U_2$. Their index functions are given by formulas, so their supports, and the intersection $\mQ = \psi_1 \cap \psi_2 = \set{(i, j) ~\middle|~ i = j, \ i + j = 4} = \set{(2, 2)}$ of these supports, are computed without storing any coordinates.}
    \label{fig:sparse-prod-inputs}
  \end{subfigure}

  \vspace{0.25em}

  \begin{subfigure}{\linewidth}
    \centering
    \includegraphics[width=0.74\linewidth]{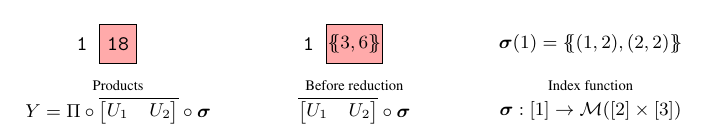}
    \caption{The product $Y$ only involves the stored values of $A_1$ and $A_2$ at the index of $\mQ$, gathered from $\stackedtensor{U_1 & U_2}$ by $\vsigma$, with $\sigma_i = \psi_i \circ \lambda$.}
    \label{fig:sparse-prod-gather}
  \end{subfigure}

  \vspace{0.25em}

  \begin{subfigure}{\linewidth}
    \centering
    \includegraphics[width=0.74\linewidth]{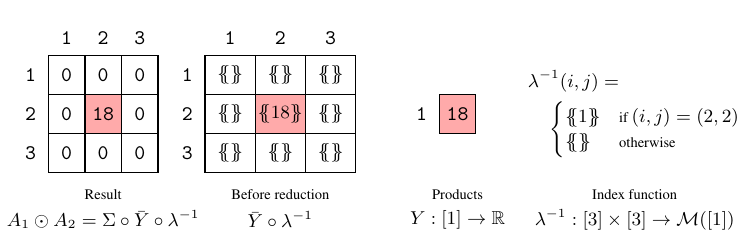}
    \caption{The product in $Y$ is scattered back by $\lambdainv$ to the only index of $\mQ$.}
    \label{fig:sparse-prod-scatter}
  \end{subfigure}

  \caption{Element-wise product $A_1 \odot A_2$ of a diagonal matrix $A_1 = \sumtensor{\psi_1}{U_1}$ and an anti-diagonal matrix $A_2 = \sumtensor{\psi_2}{U_2}$ (Theorem~\ref{thm:sparse-prod}). Only $1$ product is computed, instead of $9$. Subfigure (a) shows the inputs, (b) computes the product $Y$ from the stack of their stored values, and (c) is the right-hand side of Equation~\eqref{eq:sparse-prod}, whose underlying tensor is the result of (b). The stored values at the index of $\mQ$ are in red, and those outside of $\mQ$, which are not needed, are gray.}
  \label{fig:sparse-prod}
\end{figure}

Here, $\sigma_i$ selects the stored value of $A_i$ at the $\ell$th index of $\mQ$, so the underlying tensor of the right-hand side contains only the $L$ products that can be non-zero. The inverse $\lambdainv$ maps each $\lambda(\ell)$ to $\mset{\ell}$, and every other index to $\mset{}$, so it scatters these $L$ products back to the indices of $\mQ$. This is illustrated in Figure~\ref{fig:sparse-prod}, for $k = 2$.

As in Section~\ref{sec:examples} (Figure~\ref{fig:matmul}), an einsum between tensors $A_i : \mA_i \to \R$, for $i \in \range{k}$, whose result is indexed by $\mV$, is translated into our algebra from its subscripts. Each letter of the subscripts is a dimension of a latent index set $\mW$, the Cartesian product of the ranges of all the letters. For each operand $A_i$, the index map $\rho_i : \mW \to \mA_i$ keeps the coordinates of the letters of its subscript, in order, and the index map $\phi : \mW \to \mV$ keeps the coordinates of the letters of the output subscript, as in Figure~\ref{fig:matmul-stack} for \texttt{"mn,np->mp"}. The einsum is then the tensor
\begin{equation*}
    \einsumtensor{\phi}{A_1 \circ \rho_1 & \cdots & A_k \circ \rho_k} : \mV \to \R.
\end{equation*}
More general index maps $\rho_i$ can also be used, \eg for convolutions.
If the operands are sparse tensors, then so are the re-indexed operands $A_i \circ \rho_i = \sumtensor{\psi_i \circ \rho_i}{U_i}$, so Theorem~\ref{thm:sparse-prod} applies to the element-wise product inside the einsum. This gives the following result.

\begin{restatable}[Einsum of sparse tensors]{theorem}{thmsparseeinsum}
\label{thm:sparse-einsum}
    For each $i \in \range{k}$, let $A_i = \sumtensor{\psi_i}{U_i}: \mA_i \to \R$ be a sparse tensor, with $\psi_i: \mA_i \to \multisets{\mU_i}$ and $U_i: \mU_i \to \R$. Let $\mW$ be a latent index set, let $\rho_i : \mW \to \mA_i$, for $i \in \range{k}$, and $\phi : \mW \to \mV$ be functions, and let
    \begin{equation*}
        E = \einsumtensor{\phi}{A_1 \circ \rho_1 & \cdots & A_k \circ \rho_k} : \mV \to \R.
    \end{equation*}
    Let $\mQ = (\psi_1 \circ \rho_1) \cap \dots \cap (\psi_k \circ \rho_k) \subseteq \mW$, of size $L$, and let $\lambda: \range{L} \to \mW$ be an injective function with image $\mQ$. Then:
    \begin{equation}
        E = \sumtensor{(\phi \circ \lambda)^{-1}}{\prodtensor{\vsigma}{\stackedtensor{U_1 & \cdots & U_k}}},
        \label{eq:sparse-einsum}
    \end{equation}
    where $\vsigma$ is the stacked index function of $\sigma_i = \psi_i \circ \rho_i \circ \lambda$, for $i \in \range{k}$.
\end{restatable}

Very importantly, the product reduction in Equation~\eqref{eq:sparse-einsum} is only over the $L$ latent indices of $\mQ$, \ie the latent indices at which all operands have a stored value, instead of over the whole latent index set $\mW$. The vector $\prodtensor{\vsigma}{\stackedtensor{U_1 & \cdots & U_k}}$ thus only contains the $L$ products that can be non-zero, and $(\phi \circ \lambda)^{-1}$ maps each output index $\vv$ to the products $\ell$ such that $\phi(\lambda(\ell)) = \vv$, which are then summed.
For instance, for the product $AB$ of two sparse matrices $A: \range{M}\times\range{N}\to\R$ and $B: \range{N}\times\range{P}\to\R$, with $\phi$, $\rho_1$ and $\rho_2$ as in Figure~\ref{fig:matmul-stack}, $\mQ$ is the set of all $(m, n, p)$ such that both $A(m, n)$ and $B(n, p)$ are stored, and $(\phi \circ \lambda)^{-1}$ maps each $(m, p)$ to the products $\ell$ such that $\lambda(\ell) = (m, n, p)$ for some $n \in \range{N}$.
Dense operands are covered by taking $\psi_i = \eta$, so that $A_i = \sumtensor{\eta}{U_i} = U_i$ by Lemma~\ref{lem:reindex}, in which case $\psi_i \circ \rho_i = \eta \circ \rho_i$ has full support and does not restrict $\mQ$. For instance, the sampled dense-dense matrix multiplication (see, \eg, \cite{kjolstad2020thesis}), which multiplies a sparse matrix $A: \range{M} \times \range{P} \to \R$ element-wise by the product of two dense matrices $B: \range{M} \times \range{N} \to \R$ and $D: \range{N} \times \range{P} \to \R$, is the einsum \texttt{"mp,mn,np->mp"} of $A$, $B$ and $D$. By Theorem~\ref{thm:sparse-einsum}, if $A$ has $s$ stored values, it can be computed with only the $sN$ products at the latent indices $(m, n, p)$ such that $A(m, p)$ is stored, instead of the $MNP$ products of the dense product $BD$.

\FloatBarrier

\section{Conclusion}

We have proposed an algebra in which tensors are functions from indices to values, and in which operations on tensors are compositions of functions.

Since everything is a composition, identities between compositions can be used to rewrite expressions before evaluating them. We have shown that zero-preserving element-wise functions, element-wise products and, more generally, einsums between sparse tensors can be rewritten so that only the stored values of the operands are accessed, and only the products that can be non-zero are computed. Since the only non-trivial computations in these rewritings are on index functions, a tensor computation framework based on this algebra would not need to implement each operation for each combination of formats, which shifts the focus from defining operations on tensors to defining an algebra on index functions.

Several questions remain open. Efficient implementations of the intersection and the composition of index functions have to be designed for the most common types of index functions, such as those of the COO, compressed sparse row (CSR), diagonal and block formats. A given expression can usually be rewritten in many ways, so a cost model is needed to choose between them, as for the contraction order of einsums. Finally, generating efficient code from the rewritten expressions, \eg with the techniques of sparse tensor compilers, is left for future work. We hope that this algebra helps the development of tensor computation frameworks with much less format-specific code, and thus with support for many more sparse and structured tensor formats.

\section*{Acknowledgements}

We are deeply grateful to Pierre Quinton for his ideas that became the foundation of this work. We also thank Aart Bik for his guidance and support through the many insightful discussions we had.

\bibliographystyle{ieeetr}   %
\bibliography{references}      %

\newpage
\appendix

\section{Translation of some common PyTorch functions}
\label{app:translations}

Table~\ref{tab:translations} gives the translation of common PyTorch operations into our algebra, from the simplest to the most composite. Indices are $1$-based in our algebra and $0$-based in PyTorch, which explains the shift by one in the row of \texttt{A[:, j]}.

Some rows apply a function of several variables element-wise to several tensors. This uses the following construction.

\begin{definition}[Tupled tensor]
    Let $k \in \N$, and for each $i \in \range{k}$, let $U_i : \mU \to \S_i$ be a tensor.
    The \emph{tupled tensor}
    $\tupledtensor{U_1 & \cdots & U_k} : \mU \to \S_1 \times \dots \times \S_k$
    is defined as:
    \begin{equation}
        \tupledtensor{U_1 & \cdots& U_k} \quad:\quad \vu\in\mU \quad\mapsto\quad (U_1(\vu), \dots, U_k(\vu)).
    \end{equation}
\end{definition}

For instance, for tensors $A, B: \mU \to \R$ and a function $f: \R \times \R \to \R$, the composition $f \circ \tupledtensor{A & B}$ maps each $\vu \in \mU$ to $f(A(\vu), B(\vu))$.

\begin{longtable}{ll}
\caption{Translation of common PyTorch operations into our algebra of compositions.}
\label{tab:translations}\\
\toprule
PyTorch code & Tensor \\
\midrule
\endfirsthead
\multicolumn{2}{l}{\small\itshape Table~\ref{tab:translations} (continued).}\\
\toprule
PyTorch code & Tensor \\
\midrule
\endhead
\bottomrule
\endlastfoot
\makecell[tl]{\texttt{torch.zeros(M, N)}} &
\makecell[tl]{$\sumtensor{\psi}{U}: \range{M} \times \range{N} \to \R$\\
    \smallbullet$U: \emptyset \to \R$\\
    \smallbullet$\psi(\vv) = \mset{}$}\\
\thinrule

\makecell[tl]{\texttt{torch.ones(M, N)}} &
\makecell[tl]{$\prodtensor{\psi}{U}: \range{M} \times \range{N} \to \R$\\
    \smallbullet$U: \emptyset \to \R$\\
    \smallbullet$\psi(\vv) = \mset{}$}\\
\thinrule

\makecell[tl]{\texttt{torch.tanh(A)}\\
    \smallbullet\texttt{A.shape: (M, N)}} &
\makecell[tl]{$\tanh{} \circ A: \range{M} \times \range{N} \to \openinterval{{-1}, 1}$\\
    \smallbullet$\tanh: \R \to \openinterval{{-1}, 1}$}\\
\thinrule

\makecell[tl]{\texttt{torch.transpose(A, 0, 1)}\\
    \smallbullet\texttt{A.shape: (M, N)}} &
\makecell[tl]{$A \circ \rho: \range{N} \times \range{M} \to \R$\\
    \smallbullet$\rho(n, m) = (m, n)$}\\
\thinrule

\makecell[tl]{\texttt{A.reshape(M, N)}\\
    \smallbullet\texttt{A.shape: (M*N,)}} &
\makecell[tl]{$A \circ \rho: \range{M} \times \range{N} \to \R$\\
    \smallbullet$\rho(m, n) = N(m - 1) + n$}\\
\thinrule

\makecell[tl]{\texttt{A[:, j]}\\
    \smallbullet\texttt{A.shape: (M, N)}} &
\makecell[tl]{$A \circ \rho: \range{M} \to \R$\\
    \smallbullet$\rho(m) = (m, j + 1)$}\\
\thinrule

\makecell[tl]{\texttt{torch.diagonal(A)}\\
    \smallbullet\texttt{A.shape: (N, N)}} &
\makecell[tl]{$A \circ \rho: \range{N} \to \R$\\
    \smallbullet$\rho(n) = (n, n)$}\\
\thinrule

\makecell[tl]{\texttt{torch.sum(A, dim=0)}\\
    \smallbullet\texttt{A.shape: (M, N)}} &
\makecell[tl]{$\sumtensor{\psi}{A}: \range{N} \to \R$\\
    \smallbullet$\psi(n) = \range{M} \times \set{n}$}\\
\thinrule

\makecell[tl]{\texttt{torch.amax(A, dim=1)}\\
    \smallbullet\texttt{A.shape: (M, N)}} &
\makecell[tl]{$\maxtensor{\psi}{A}: \range{M} \to \R$\\
    \smallbullet$\psi(m) = \set{m} \times \range{N}$}\\
\thinrule

\makecell[tl]{\texttt{torch.mean(A, dim=1)}\\
    \smallbullet\texttt{A.shape: (M, N)}} &
\makecell[tl]{$g \circ \sumtensor{\psi}{A}: \range{M} \to \R$\\
    \smallbullet$\psi(m) = \set{m} \times \range{N}$\\
    \smallbullet$g(x) = x / N$}\\
\thinrule

\makecell[tl]{\texttt{torch.cumsum(A, dim=0)}\\
    \smallbullet\texttt{A.shape: (N,)}} &
\makecell[tl]{$\sumtensor{\psi}{A}: \range{N} \to \R$\\
    \smallbullet$\psi(n) = \mset{1, \dots, n}$}\\
\thinrule

\makecell[tl]{\texttt{torch.trace(A)}\\
    \smallbullet\texttt{A.shape: (N, N)}} &
\makecell[tl]{$\sumtensor{\psi}{A}: \range{1} \to \R$\\
    \smallbullet$\psi(1) = \set{(n, n) ~|~ n \in \range{N}}$}\\
\thinrule

\makecell[tl]{\texttt{torch.diag(A)}\\
    \smallbullet\texttt{A.shape: (N,)}} &
\makecell[tl]{$\sumtensor{\psi}{A}: \range{N} \times \range{N} \to \R$\\
    \smallbullet$\psi(i, j) = \mset{i}$ if $i = j$; $\mset{}$ otherwise}\\
\thinrule

\makecell[tl]{\texttt{torch.mul(A, B)}\\
    \smallbullet\texttt{A.shape: (M, N)}\\
    \smallbullet\texttt{B.shape: (M, N)}} & $\prodtensor{\vid}{\stackedtensor{A & B}}: \range{M} \times \range{N} \to \R$ \\
\thinrule

\makecell[tl]{\texttt{torch.float\_power(A, B)} \\
    \smallbullet\texttt{A.shape: (M, N)} \\
    \smallbullet\texttt{B.shape: (M, N)}} &
\makecell[tl]{$f \circ \tupledtensor{A & B}: \range{M} \times \range{N} \to \R$\\
    \smallbullet $f(a, b) = a^b$}\\
\thinrule

\makecell[tl]{\texttt{torch.where(K, A, B)} \\
    \smallbullet\texttt{A.shape: (M, N)} \\
    \smallbullet\texttt{B.shape: (M, N)} \\
    \smallbullet\texttt{K.shape: (M, N)}} & 
\makecell[tl]{$\sumtensor{\vxi}{\stackedtensor{A & B}}: \range{M} \times \range{N} \to \R$\\
    \smallbullet$\xi_1(\vv) = \mset{\vv}$ if $K(\vv) = 1$; $\mset{}$ otherwise.\\
    \smallbullet$\xi_2(\vv)=\mset{\vv}$ if $K(\vv) = 0$; $\mset{}$ otherwise.}\\
\thinrule

\makecell[tl]{\texttt{torch.outer(A, B)}\\
    \smallbullet\texttt{A.shape: (M,)}\\
    \smallbullet\texttt{B.shape: (N,)}} &
\makecell[tl]{$\prodtensor{\vid}{\stackedtensor{A \circ \rho_1 & B \circ \rho_2}}: \range{M} \times \range{N} \to \R$\\
    \smallbullet$\rho_1(m, n) = m$\\
    \smallbullet$\rho_2(m, n) = n$}\\
\thinrule

\makecell[tl]{\texttt{torch.softmax(A, dim=0)}\\
    \smallbullet\texttt{A.shape: (N,)}} &
\makecell[tl]{$f \circ \tupledtensor{E & \sumtensor{\psi}{E}}: \range{N} \to \R$\\
    \smallbullet$E = \exp \circ A$\\
    \smallbullet$\psi(n) = \mset{1, \dots, N}$\\
    \smallbullet$f(a, b) = a / b$}\\
\thinrule

\makecell[tl]{\texttt{torch.matmul(A, B)}\\
    \smallbullet\texttt{A.shape: (M, N)}\\
    \smallbullet\texttt{B.shape: (N, P)}} & 
\makecell[tl]{$\sumtensor{\phi^{-1}}{\prodtensor{\vid}{\stackedtensor{A \circ \rho_1 & B \circ \rho_2}}}$\\
    \quad$: \range{M} \times \range{P} \to \R$\\
    \smallbullet$\rho_1(m, n, p) = (m, n)$\\
    \smallbullet$\rho_2(m, n, p) = (n, p)$\\
    \smallbullet$\phi(m, n, p) = (m, p)$, so\\
    \smallbullet $\phi^{-1}(m, p) = \set{m} \times \range{N} \times \set{p}$} \\
\thinrule

\makecell[tl]{\texttt{torch.einsum(}\\
    \texttt{\phantom{to}"mnp,pnq->mq", A, B)}\\
    \smallbullet\texttt{A.shape: (M, N, P)}\\
    \smallbullet\texttt{B.shape: (P, N, Q)}} & 
\makecell[tl]{$\sumtensor{\phi^{-1}}{\prodtensor{\vid}{\stackedtensor{A \circ \rho_1 & B \circ \rho_2}}}$\\
    \quad$: \range{M} \times \range{Q} \to \R$\\
    \smallbullet$\rho_1(m, n, p, q) = (m, n, p)$\\
    \smallbullet$\rho_2(m, n, p, q) = (p, n, q)$\\
    \smallbullet$\phi(m, n, p, q) = (m, q)$, so\\
    \smallbullet $\phi^{-1}(m, q) = \set{m} \times \range{N} \times \range{P} \times \set{q}$}\\
\end{longtable}

\section{Proofs of Section~\ref{sec:rewriting}}
\label{app:proofs}

We first give general identities between compositions as lemmas (Appendix~\ref{app:lemmas}), from which we prove the theorems of Section~\ref{sec:rewriting} (Appendix~\ref{app:thm-proofs}).

\paragraph{Notation.}
Recall that $\eta : \mathcal{X} \to \multisets{\mathcal{X}}$ denotes the singleton map of Section~\ref{sec:theory}, defined as $\eta(x) = \mset{x}$. For any set $\mathcal{X}$, we denote by $\mu : \multisets{\multisets{\mathcal{X}}} \to \multisets{\mathcal{X}}$ the \emph{flattening} of multisets of multisets, defined as $\mu\left(\mset{X_1, \dots, X_n}\right) = X_1 \uplus \dots \uplus X_n$.
We also call \emph{support} of a tensor $B : \mV \to \R$ the set $\supp B = \set{\vv \in \mV ~\middle|~ B(\vv) \neq 0}$.

\subsection{Lemmas}
\label{app:lemmas}

\begin{lemma}[Identities on multisets]
\label{lem:multisets}
    Let $f : \mV \to \mU$ and $g : \mU \to \mW$ be functions. Then:
    \begin{align}
        \lift{g \circ f} &= \lift{g} \circ \lift{f}, \label{eq:rw-functor}\\
        \lift{f} \circ \eta &= \eta \circ f, \label{eq:rw-eta}\\
        \mu \circ \bar{\bar{f}} &= \lift{f} \circ \mu, \label{eq:rw-mu}\\
        (g \circ f)^{-1} &= \mu \circ \lift{f^{-1}} \circ g^{-1}. \label{eq:rw-inv}
    \end{align}
\end{lemma}

\begin{proof}
    Equations~\eqref{eq:rw-functor} to~\eqref{eq:rw-mu} follow directly from the definitions: extension to multisets commutes with composition, applying $f$ to the element of a singleton gives the singleton of the image, and applying $f$ to every element of every multiset and then flattening is the same as flattening and then applying $f$ to every element.
    For Equation~\eqref{eq:rw-inv}, both sides map each $\vw \in \mW$ to the multiset of all $\vv \in \mV$ such that $g(f(\vv)) = \vw$, which the right-hand side groups by the value of $f(\vv) \in g^{-1}(\vw)$.
\end{proof}

\begin{lemma}[Reductions]
\label{lem:reductions}
    Let $\Omega$ be a reduction on $\S$. Then:
    \begin{align}
        \Omega \circ \eta &= \id, \label{eq:rw-unit}\\
        \Omega \circ \lift{\Omega} &= \Omega \circ \mu. \label{eq:rw-assoc}
    \end{align}
\end{lemma}

\begin{proof}
    Equation~\eqref{eq:rw-unit} is the second property of reductions. Equation~\eqref{eq:rw-assoc} says that reducing the reductions of several multisets amounts to reducing their sum. It follows from the third property (associativity), by induction on the number $n$ of multisets: $\Omega(\mset{\Omega(X_1), \dots, \Omega(X_n)}) = \Omega(\mset{\Omega(\mset{\Omega(X_1), \dots, \Omega(X_{n-1})}), \Omega(X_n)}) = \Omega(\mset{\Omega(X_1 \uplus \dots \uplus X_{n-1}), \Omega(X_n)}) = \Omega(X_1 \uplus \dots \uplus X_n)$, the cases $n = 0$ and $n = 1$ following from the first two properties of reductions.
\end{proof}

In the rest of this section, unless stated otherwise, $U : \mU \to \S$ is a tensor and $\Omega$ is a reduction on $\S$.

\begin{lemma}[Re-indexing as a reduction]
\label{lem:reindex}
    For any function $f : \mV \to \mU$:
    \begin{equation}
        \omegatensor{\eta \circ f}{U} = U \circ f.
        \label{eq:rw-reindex}
    \end{equation}
    Consequently, if $\psi : \mV \to \multisets{\mU}$ maps every index to a singleton, then, for any reduction $\Omega'$ on $\S$:
    \begin{equation}
        \omegatensor{\psi}{U} = \optensor{\psi}{U}{\Omega'}.
        \label{eq:rw-change}
    \end{equation}
\end{lemma}

\begin{proof}
    By Equations~\eqref{eq:rw-eta} and~\eqref{eq:rw-unit}, $\Omega \circ \bar{U} \circ \eta \circ f = \Omega \circ \eta \circ U \circ f = U \circ f$. For Equation~\eqref{eq:rw-change}, $\psi = \eta \circ f$, where $f(\vv)$ is the element of $\psi(\vv)$, so both sides are equal to $U \circ f$ by Equation~\eqref{eq:rw-reindex}.
\end{proof}

A re-indexing is thus a reduction tensor whose index function only returns singletons. In particular, with $f = \id$, any tensor is the reduction tensor of itself: $U = \omegatensor{\eta}{U}$.

\begin{lemma}[Re-indexing reduction tensors]
\label{lem:pull-push}
    For any index function $\psi : \mV \to \multisets{\mU}$ and any function $\rho : \mW \to \mV$:
    \begin{equation}
        \left(\omegatensor{\psi}{U}\right) \circ \rho = \omegatensor{\psi \circ \rho}{U}.
        \label{eq:rw-pull}
    \end{equation}
    For any function $f : \mU' \to \mU$ and any index function $\chi : \mV \to \multisets{\mU'}$:
    \begin{equation}
        \omegatensor{\chi}{U \circ f} = \omegatensor{\lift{f} \circ \chi}{U}.
        \label{eq:rw-push}
    \end{equation}
\end{lemma}

\begin{proof}
    Equation~\eqref{eq:rw-pull} follows from the associativity of the composition, and Equation~\eqref{eq:rw-push} from $\lift{U \circ f} = \bar{U} \circ \lift{f}$, by Equation~\eqref{eq:rw-functor}.
\end{proof}

Re-indexing a reduction tensor thus amounts to re-indexing its index function, and re-indexing its underlying tensor amounts to applying the re-indexing to each element of the multisets returned by its index function.

\begin{lemma}[Nested reductions]
\label{lem:nested}
    For any index functions $\alpha : \mW \to \multisets{\mU}$ and $\beta : \mV \to \multisets{\mW}$:
    \begin{equation}
        \omegatensor{\beta}{\omegatensor{\alpha}{U}} = \omegatensor{\mu \circ \lift{\alpha} \circ \beta}{U}.
        \label{eq:nested}
    \end{equation}
\end{lemma}

\begin{proof}
    By Equations~\eqref{eq:rw-functor}, \eqref{eq:rw-assoc} and~\eqref{eq:rw-mu}, successively:
    \begin{equation*}
        \omegatensor{\beta}{\omegatensor{\alpha}{U}}
        = \Omega \circ \lift{\Omega} \circ \bar{\bar{U}} \circ \lift{\alpha} \circ \beta
        = \Omega \circ \mu \circ \bar{\bar{U}} \circ \lift{\alpha} \circ \beta
        = \Omega \circ \bar{U} \circ \mu \circ \lift{\alpha} \circ \beta.
        \qedhere
    \end{equation*}
\end{proof}

The index function $\mu \circ \lift{\alpha} \circ \beta$ maps each $\vv$ to the multiset of all $\vu \in \alpha(\vw)$ for $\vw \in \beta(\vv)$. Two successive reductions with the same $\Omega$ thus collapse into a single one. Note that this does not hold for two different reductions, such as a sum of products.

\begin{lemma}[Homomorphisms]
\label{lem:hom}
    Let $\Omega'$ be a reduction on some set $\S'$, let $g : \S \to \S'$ be a function such that $g \circ \Omega = \Omega' \circ \lift{g}$, and let $\psi : \mV \to \multisets{\mU}$ be an index function. Then:
    \begin{equation}
        g \circ \omegatensor{\psi}{U} = \optensor{\psi}{g \circ U}{\Omega'}.
        \label{eq:rw-hom}
    \end{equation}
    If $\psi$ maps every index to a multiset containing at most one element, the same holds as soon as $g(e) = e'$, where $e$ and $e'$ are the neutral elements of $\Omega$ and $\Omega'$.
\end{lemma}

\begin{proof}
    By Equation~\eqref{eq:rw-functor}, $g \circ \Omega \circ \bar{U} \circ \psi = \Omega' \circ \lift{g} \circ \bar{U} \circ \psi = \Omega' \circ \lift{g \circ U} \circ \psi$. If $\psi$ maps every index to a multiset containing at most one element, only the equality $g \circ \Omega = \Omega' \circ \lift{g}$ on such multisets is needed. On the empty multiset, it reduces to $g(e) = e'$, and on a singleton $\mset{x}$, both sides are equal to $g(x)$.
\end{proof}

For instance, $g(x) = c x$ satisfies the condition with $\Omega = \Omega' = \Sigma$, $\exp$ satisfies it with $\Omega = \Sigma$ and $\Omega' = \Pi$, and any non-decreasing $g$ such that $g(-\infty) = -\infty$ satisfies it with $\Omega = \Omega' = \vee$.

\begin{lemma}[Splitting a reduction over a stack]
\label{lem:split}
    For tensors $A_i : \mA_i \to \S$ and index functions $\psi_i : \mV \to \multisets{\mA_i}$:
    \begin{equation}
        \omegatensor{\vpsi}{\stackedtensor{A_1 & \cdots & A_k}} = \omegatensor{\vid}{\stackedtensor{\omegatensor{\psi_1}{A_1} & \cdots & \omegatensor{\psi_k}{A_k}}}.
        \label{eq:rw-split}
    \end{equation}
\end{lemma}

\begin{proof}
    Let $\vv \in \mV$, and let $X_i = \bar{A_i}(\psi_i(\vv))$ for each $i \in \range{k}$. By definition of $\vpsi$, the left-hand side maps $\vv$ to $\Omega(X_1 \uplus \dots \uplus X_k)$. Since $\vid(\vv) = \mset{(1, \vv), \dots, (k, \vv)}$, the right-hand side maps $\vv$ to $\Omega(\mset{\Omega(X_1), \dots, \Omega(X_k)})$. These are equal by Equation~\eqref{eq:rw-assoc}, applied to $\mset{X_1, \dots, X_k}$.
\end{proof}

Reducing a stack with a stacked index function thus amounts to reducing each tensor with its own index function, and then combining the results element-wise with the same reduction. Read from right to left, Equation~\eqref{eq:rw-split} says that an element-wise reduction of reduction tensors, with the same reduction $\Omega$, is a single reduction tensor over the stack of their underlying tensors.

\begin{lemma}[Moving index maps]
\label{lem:move}
    For tensors $A_i : \mA_i \to \S$ and functions $\rho_i : \mB \to \mA_i$:
    \begin{equation}
        \omegatensor{\vid}{\stackedtensor{A_1 \circ \rho_1 & \cdots & A_k \circ \rho_k}} = \omegatensor{\vrho}{\stackedtensor{A_1 & \cdots & A_k}}.
        \label{eq:move-stack}
    \end{equation}
    In particular, element-wise operations commute with re-indexing: for tensors $B_i : \mV \to \S$ and any function $\rho : \mW \to \mV$:
    \begin{equation}
        \left(\omegatensor{\vid}{\stackedtensor{B_1 & \cdots & B_k}}\right) \circ \rho = \omegatensor{\vid}{\stackedtensor{B_1 \circ \rho & \cdots & B_k \circ \rho}}.
        \label{eq:rw-ew-reindex}
    \end{equation}
\end{lemma}

\begin{proof}
    Since $\vrho$ is the stacked index function of the $\eta \circ \rho_i$, Equation~\eqref{eq:rw-split} with $\psi_i = \eta \circ \rho_i$ shows that the right-hand side of Equation~\eqref{eq:move-stack} is equal to $\omegatensor{\vid}{\stackedtensor{\omegatensor{\eta \circ \rho_1}{A_1} & \cdots & \omegatensor{\eta \circ \rho_k}{A_k}}}$, and $\omegatensor{\eta \circ \rho_i}{A_i} = A_i \circ \rho_i$ by Equation~\eqref{eq:rw-reindex}.
    For Equation~\eqref{eq:rw-ew-reindex}, by Equation~\eqref{eq:rw-pull}, the left-hand side is $\omegatensor{\vid \circ \rho}{\stackedtensor{B_1 & \cdots & B_k}}$, and $\vid \circ \rho$ is the stacked index function of the functions $\rho_i = \rho$, so Equation~\eqref{eq:move-stack} gives the right-hand side.
\end{proof}

\begin{lemma}[Supports]
\label{lem:supports}
    For any index function $\psi : \mV \to \multisets{\mU}$ and tensors $U : \mU \to \R$ and $B_1, \dots, B_k : \mV \to \R$:
    \begin{align}
        \supp \left(\sumtensor{\psi}{U}\right) &\subseteq \supp \psi, \label{eq:rw-supp-red}\\
        \supp \left(\prodtensor{\vid}{\stackedtensor{B_1 & \cdots & B_k}}\right) &= \supp B_1 \cap \dots \cap \supp B_k. \label{eq:rw-supp-prod}
    \end{align}
\end{lemma}

\begin{proof}
    Equation~\eqref{eq:rw-supp-red} holds because a sum tensor is zero outside of the support of its index function, and Equation~\eqref{eq:rw-supp-prod} because a product of real numbers is zero if and only if one of its factors is zero.
\end{proof}

\begin{lemma}[Compaction]
\label{lem:compact}
    Let $\mQ \subseteq \mV$ be a set of size $L$, and let $\lambda: \range{L} \to \mV$ be an injective function with image $\mQ$. For any tensor $B : \mV \to \R$ such that $\supp B \subseteq \mQ$:
    \begin{equation}
        B = \sumtensor{\lambdainv}{B \circ \lambda}.
        \label{eq:rw-compact}
    \end{equation}
\end{lemma}

\begin{proof}
    At $\lambda(\ell)$, the right-hand side is the sum of $\mset{B(\lambda(\ell))}$, \ie $B(\lambda(\ell))$, and outside of $\mQ$, both sides are zero.
\end{proof}

The vector $B \circ \lambda : \range{L} \to \R$ gathers the $L$ values of $B$ that can be non-zero, and $\lambdainv$ scatters them back. This holds for any reduction, if $0$ is replaced by its neutral element. The proof of Theorem~\ref{thm:sparse-prod} follows this pattern: we bound the support of the result with Lemma~\ref{lem:supports}, compact it with Lemma~\ref{lem:compact}, and simplify $Z \circ \lambda$ with the other lemmas.

\subsection{Proofs of the main results}
\label{app:thm-proofs}

\thmsparseew*

\begin{proof}
    In general, $f \circ \Sigma \neq \Sigma \circ \lift{f}$ (\eg for $f(x) = x^2$), but they are equal on multisets containing at most one element: on $\mset{}$, both give $f(0) = 0$, and on $\mset{x}$, both give $f(x)$. Since $\psi$ maps every index to such a multiset, the second part of Lemma~\ref{lem:hom}, with $\Omega = \Omega' = \Sigma$ and $g = f$, gives Equation~\eqref{eq:sparse-ew}.
\end{proof}

\thmsparseprod*

\begin{proof}
    Let $Z$ be the left-hand side of Equation~\eqref{eq:sparse-prod}. By Equations~\eqref{eq:rw-supp-prod} and~\eqref{eq:rw-supp-red}, $\supp Z = \supp A_1 \cap \dots \cap \supp A_k \subseteq \supp \psi_1 \cap \dots \cap \supp \psi_k = \mQ$, so $Z = \sumtensor{\lambdainv}{Z \circ \lambda}$ by Equation~\eqref{eq:rw-compact}. Since $\lambda(\ell) \in \supp \psi_i$ for all $\ell$, and $\psi_i$ maps every index to a multiset containing at most one element, each $\sigma_i$ maps every $\ell$ to a singleton. Then, by Equations~\eqref{eq:rw-ew-reindex}, \eqref{eq:rw-pull}, \eqref{eq:rw-change} and~\eqref{eq:rw-split}, successively:
    \begin{align*}
        Z \circ \lambda
        &= \prodtensor{\vid}{\stackedtensor{A_1 \circ \lambda & \cdots & A_k \circ \lambda}}
        = \prodtensor{\vid}{\stackedtensor{\sumtensor{\sigma_1}{U_1} & \cdots & \sumtensor{\sigma_k}{U_k}}}\\
        &= \prodtensor{\vid}{\stackedtensor{\prodtensor{\sigma_1}{U_1} & \cdots & \prodtensor{\sigma_k}{U_k}}}
        = \prodtensor{\vsigma}{\stackedtensor{U_1 & \cdots & U_k}}.
        \qedhere
    \end{align*}
\end{proof}

\thmsparseeinsum*

\begin{proof}
    By Equation~\eqref{eq:rw-pull}, each $A_i \circ \rho_i = \sumtensor{\psi_i \circ \rho_i}{U_i}$ is a sparse tensor on the latent index set $\mW$, since $\psi_i \circ \rho_i$ maps every index to a multiset containing at most one element. Theorem~\ref{thm:sparse-prod}, applied to these tensors, followed by Equations~\eqref{eq:nested} and~\eqref{eq:rw-inv}, gives, with $Y = \prodtensor{\vsigma}{\stackedtensor{U_1 & \cdots & U_k}}$:
    \begin{align*}
        E &= \sumtensor{\phi^{-1}}{\prodtensor{\vid}{\stackedtensor{A_1 \circ \rho_1 & \cdots & A_k \circ \rho_k}}}
        = \sumtensor{\phi^{-1}}{\sumtensor{\lambdainv}{Y}}\\
        &= \sumtensor{\mu \circ \lift{\lambda^{-1}} \circ \phi^{-1}}{Y}
        = \sumtensor{(\phi \circ \lambda)^{-1}}{Y}.
        \qedhere
    \end{align*}
\end{proof}

\end{document}